\pdfoutput=1

\documentclass[11pt]{article}

\usepackage{acl}

\usepackage{times}
\usepackage{latexsym}

\usepackage{multirow,array,float,graphicx}

\usepackage{amsmath,amssymb}

\usepackage[T1]{fontenc}

\usepackage[utf8]{inputenc}

\usepackage{microtype}

\usepackage{booktabs}

\usepackage{xcolor}

\usepackage{comment}
\usepackage{soul}

\title{Modelling Commonsense Properties using Pre-Trained Bi-Encoders}

\author{Amit Gajbhiye$^{*}$, Luis Espinosa-Anke$^{*\diamondsuit}$, Steven Schockaert$^{*}$ \\
         $^{*}$CardiffNLP, Cardiff University, United Kingdom \\ $^{\diamondsuit}$AMPLYFI, United Kingdom\\ 
         \texttt{\{gajbhiyea, espinosa-ankel, schockaerts1\}@cardiff.ac.uk}}

\begin{document}
\maketitle

\begin{abstract}
Grasping the commonsense properties of everyday concepts is an important prerequisite to language understanding. While contextualised language models are reportedly capable of predicting such commonsense properties with human-level accuracy, we argue that such results have been inflated because of the high similarity between training and test concepts. This means that models which capture concept similarity can perform well, even if they do not capture any knowledge of the commonsense properties themselves. In settings where there is no overlap between the properties that are considered during training and testing, we find that the empirical performance of standard language models drops dramatically. To address this, we study the possibility of fine-tuning language models to explicitly model concepts and their properties. In particular, we train separate concept and property encoders on two types of readily available data: extracted hyponym-hypernym pairs and generic sentences. Our experimental\footnote{Code and datasets are available at \url{https://github.com/amitgajbhiye/biencoder\_concept\_property}} results show that the resulting encoders allow us to predict commonsense properties with much higher accuracy than is possible by directly fine-tuning language models. We also present experimental results for the related task of unsupervised hypernym discovery.
\end{abstract}

\section{Introduction}
Pre-trained language models \cite{devlin-etal-2019-bert} have been found to capture a surprisingly rich amount of knowledge about the world \cite{DBLP:conf/emnlp/PetroniRRLBWM19}. Focusing on commonsense knowledge, \citet{DBLP:conf/cogsci/ForbesHC19} used BERT to predict whether a given concept (e.g.\ \emph{teddy bear}) satisfies a given commonsense property (e.g.\ \emph{is dangerous}). To this end, they convert the input into a simple sentence (e.g.\ ``\emph{A teddy bear is dangerous}'') and treat the task as a standard sentence classification task. Remarkably, they found this approach to surpass human performance. 
\citet{shwartz-choi-2020-neural} moreover found that language models can, to some extent, capture commonsense properties that are rarely expressed in text, thus mitigating the issue of reporting bias that has traditionally plagued initiatives for learning commonsense knowledge from text \cite{DBLP:conf/cikm/GordonD13}. 

Despite these encouraging signs, however, modelling commonsense properties remains highly challenging. A key concern is that language models are typically fine-tuned on a training set that contains the same properties as those in the test set. For instance, the test data from \citet{DBLP:conf/cogsci/ForbesHC19} includes the question whether \emph{peach} has the property \emph{eaten in summer}, while the training data asserts that \emph{apple}, \emph{banana}, \emph{orange} and \emph{pear} all have this property. To do well on this task, the model does not actually need to capture the knowledge that peaches are eaten in summer; it is sufficient to capture that \emph{peach} is similar to \emph{apple}, \emph{banana}, \emph{orange} and \emph{pear}. For this reason, we propose new training-test splits, which ensure that the properties occurring in the test data do not occur in the training data. Our experiments show that the ability of language models to predict commonsense properties drops dramatically in this setting. 

\begin{table*}
\centering
\footnotesize
\begin{tabular}{lp{160pt}p{120pt}p{120pt}}
\toprule
& \multicolumn{1}{c}{\textbf{ConceptNet}} & \multicolumn{1}{c}{\textbf{COMET-2020}} & \multicolumn{1}{c}{\textbf{Ascent++}}\\
\midrule
 \parbox[t]{2mm}{\multirow{3}{*}{\rotatebox[origin=c]{90}{\textbf{banana}}}} & yellow, good to eat & one of the main ingredients, eaten as a snack, one of many fruits, found in garden, black & rich, ripe, yellow, green, brown, sweet, great, black, useful, safe, delicious, healthy, nutricious, ...\\[0.5em]
\midrule
\parbox[t]{2mm}{\multirow{3}{*}{\rotatebox[origin=c]{90}{\textbf{lion}}}} & a feline & found in jungle, one of many animals, one of many species, two legs, very large & free, extinct, hungry, close, unique, active, nocturnal, old, dangerous, great, happy, right, ...\\[0.5em]
\midrule
\parbox[t]{2mm}{\multirow{3}{*}{\rotatebox[origin=c]{90}{\textbf{airplane}}}}& good for quickly travelling long distances & flying, air travel, flying machine, very small, flight & heavy, new, important, white, safe, unique, full, larger, clean, slow, low, unstable, electric, ...\\[2.5em]
\bottomrule
\end{tabular}
\caption{Properties of some example concepts, according to three  commonsense knowledge resources.\label{tabPropertiesIntro}}
\end{table*}

Our aim is to develop a strategy for modelling the commonsense properties of concepts. Given the limitations that arise when language models are used directly, a natural approach is to pre-train a language model on some kind of auxiliary data. Unfortunately, resources encoding the commonsense properties of concepts tend to be prohibitively noisy. To illustrate this point, Table \ref{tabPropertiesIntro} lists the properties of some everyday concepts according to three well-known resources: ConceptNet \cite{DBLP:conf/aaai/SpeerCH17}, which is a large commonsense knowledge graph, COMET-2020\footnote{We used the demo at \url{https://mosaickg.apps.allenai.org/model_comet2020_entities}.} \cite{DBLP:conf/aaai/HwangBBDSBC21}, which predicts triples using a generative language model that was trained on several commonsense knowledge graphs, and Ascent++ \cite{DBLP:journals/corr/abs-2112-04596}, which is a commonsense knowledge base that was extracted from web text. Given the noisy nature of such resources, we rely on a database with hypernyms instead. The underlying intuition is that hypernyms can be extracted from text relatively easily, while fine-grained hypernyms often implicitly describe commonsense properties. For instance, Microsoft Concept Graph \cite{DBLP:journals/dint/JiWSZWY19} lists \emph{potassium rich food} as a hypernym of \emph{banana} and 
\emph{large and dangerous carnivore} as a hypernym of \emph{lion}.
We also experiment with GenericsKB \cite{DBLP:journals/corr/abs-2005-00660}, a large collection of generic sentences (e.g.\ ``\emph{Coffee contains minerals and antioxidants which help prevent diabetes}''), to obtain concept-property pairs for pre-training.
Given such pre-training data, we then train a concept encoder $\Phi_\mathsf{con}$ and a property encoder $\Phi_\mathsf{prop}$ such that $\sigma(\Phi_\mathsf{con}(c)\cdot \Phi_\mathsf{prop}(p))$ indicates the probability that concept $c$ has property $p$. 
In summary, our main contributions are as follows: (i) we propose a new evaluation setting which is more realistic than the standard benchmarks for predicting commonsense properties; (ii) we analyse the potential of hypernymy datasets and generic sentences to act as pre-training data; and (iii) we develop a simple but effective bi-encoder architecture for modelling commonsense properties.



\section{Related Work}
Several authors have analysed the extent to which language models such as BERT capture commonsense knowledge. As already mentioned, \citet{DBLP:conf/cogsci/ForbesHC19} evaluated the ability of BERT to predict commonsense properties from the McRae dataset \cite{mcrae2005semantic}, which we also use in our experiments. 
The same dataset was used by \citet{DBLP:conf/cogsci/WeirPD20} to analyse whether BERT-based language models could generate concept names from their associated properties; e.g.\ given the input ``\emph{A $\langle \textit{mask} \rangle$ has fur, is big, and has claws}'', the model is expected to predict that $\langle \textit{mask} \rangle$ corresponds to the word \emph{bear}. Conversely, \citet{apidianaki-gari-soler-2021-dolphins} considered the problem of generating adjectival properties from prompts such as ``mittens are generally $\langle\textit{mask}\rangle$''. Note that the latter two works evaluated pre-trained models directly, without fine-tuning, whereas the experiments \citet{DBLP:conf/cogsci/ForbesHC19} involved fine-tuning the language model on a task-specific training set first. When the main motivation is to probe the abilities of language models, avoiding fine-tuning has the advantage that any observed abilities reflect what is captured by the pre-trained language model itself, rather than learned during the fine-tuning phase. 
However, \citet{DBLP:journals/corr/abs-2111-00607} argue that the extent to which pre-trained language models capture commonsense knowledge is limited, suggesting that some form of fine-tuning is essential in practice. Interestingly, this remains the case for large language models. For instance, their model had 7 billion parameters, while \citet{DBLP:journals/corr/abs-2110-07178} report that the predictions from GPT-3 \cite{brown2020language} had to be filtered by a so-called critic model when distilling a commonsense knowledge graph.

The strategy taken by COMET \cite{bosselut-etal-2019-comet} is to fine-tune a GPT model \cite{GPT} on triples from commonsense knowledge graphs. Being based on an autoregressive language model, COMET can be used to predict concepts that take the form of short phrases, which is often needed when reasoning about events (e.g.\ to express motivations or effects). However, as illustrated in Table \ref{tabPropertiesIntro}, COMET is less suitable for modelling the commonsense properties of concepts. 
Other approaches have focused on improving the commonsense reasoning abilities of general purpose language models. For instance, \citet{DBLP:conf/iclr/ZhouLSL021} introduce a self-supervised pre-training tasks to encourage language models to better capture the commonsense relations between everyday concepts.

A final line of related work concerns the modelling of hypernymy. Several authors have proposed specialised embedding models for this task \cite{DBLP:journals/corr/abs-2106-14361,le-etal-2019-inferring}. Most relevant to our work, \citet{takeoka-etal-2021-low} fine-tune a BERT-based language model to predict the validity of a concept--hypernym pair. Inspired by the effectiveness of Hearst patterns \cite{hearst-1992-automatic}, they use prompts of the form ``[HYPERNYM] such as [CONCEPT]'' (and similar for other Hearst patterns).  The extent to which the pre-trained BERT model captures hypnernymy has also been studied. For instance, \citet{hanna-marecek-2021-analyzing} use prompts where the prediction of the $\langle\textit{mask}\rangle$ token can be interpreted as the prediction of a hypernym, to avoid the need for fine-tuning the model. 



\section{Methodology}\label{secMethodology}
Let a set of concept--property pairs $\mathcal{K}$ be given, where $(c,p)\in\mathcal{K}$ means that concept $c$ is asserted to have the property $p$. We write $\mathcal{C}$ and $\mathcal{P}$ for the sets of concepts and properties in $\mathcal{K}$, i.e.\ $\mathcal{C}=\{c \mid (c,p)\in \mathcal{K}\}$ and $\mathcal{P}=\{p \mid (c,p)\in \mathcal{K}\}$. We use the term ``property'' in a broad sense, covering both semantic attributes, as in the pair $(\emph{banana},\emph{sweet})$, and hypernyms, as in the pair $(\emph{banana},\emph{fruit})$. This is motivated by the fact that hypernyms often encode knowledge about semantic attributes, as in the pair  $(\emph{banana},\emph{sweet fruit})$. In particular, our hypothesis is that, by treating hypernyms and semantic attributes in a unified way, we can pre-train a model on readily available hypernym datasets and use it to predict semantic attributes.

We want to train a model that can predict for a given pair $(c,p)$ whether $c$ has property $p$. Two general strategies can be pursued when developing such models. The first strategy is to use a so-called cross-encoder, which amounts to fine-tuning a single language model to predict whether a given input $(c,p)$ represents a valid pair or not. The second strategy is to use a so-called bi-encoder, which amounts to the idea that $c$ and $p$ are separately encoded, with the resulting vectors then being used to predict whether $(c,p)$ is a valid pair. In this paper, we pursue the latter strategy. This is primarily motivated by the fact that the concept and property encoders enable a wider range of applications. A cross-encoder can only be used to predict whether a given pair $(c,p)$ is valid or not. In contrast, a bi-encoder model can also be used to efficiently find the properties $p$ of a given concept $c$. Moreover, the resulting concept and property embeddings may themselves be useful as static representations of word meaning, e.g.\ as label embeddings for zero-shot or few-shot learning \cite{DBLP:conf/nips/SocherGMN13,ma-etal-2016-label,DBLP:conf/nips/XingROP19,DBLP:conf/cvpr/Li0LFLW20,DBLP:conf/mir/YanBWJS21}. Finally, bi-encoders can be trained more efficiently than cross-encoders. 

\paragraph{Datasets}
To train our model, we need a large set of concept--property pairs $\mathcal{K}$. Unfortunately, high-quality knowledge of this kind is not readily available. Part of the underlying issue is that properties of concepts are rarely explicitly stated in text, which is why directly using information extraction techniques is not straightforward. However, initiatives for extracting hypernyms from text have been much more successful, starting with the seminal work of \citet{hearst-1992-automatic}. A key observation is that fine-grained hypernyms often express commonsense properties, typically as a mechanism for refining hypernyms that would otherwise be too broad. For instance, Microsoft Concept Graph \cite{DBLP:journals/dint/JiWSZWY19} lists \emph{vitamin C rich food} as a hypernym of \emph{strawberry}, as a refinement of the more general hypernym \emph{food}. By pre-training our model on concept--hypernym pairs, we may thus expect it to learn about commonsense properties as well. To directly test this hypothesis, we use a set of such concept--hypernym pairs as our pre-training set $\mathcal{K}$. Specifically, we collect the 100K concept--hypernym pairs from Microsoft Concept Graph\footnote{\url{https://concept.research.microsoft.com/Home/Download}} with the highest confidence score\footnote{Specifically, we used those pairs maximising the \emph{Relations} frequency.} 
We will refer to this dataset as \textsc{MSCG}. 

As a second strategy, we attempt to convert the MSCG dataset into a set of concept--property pairs. To this end, we look for pairs $(c,h_1)$ and $(c,h_2)$ where $h_2$ is a suffix of $h_1$. Specifically, if $h_1 = m h_2$ and $m$ is an adjectival phrase, then we assume that $m$ describes a property of $c$. For instance, MSCG contains the pairs (\emph{strawberry}, \emph{vitamin C rich food}) and (\emph{strawberry}, \emph{food}). Based on this, we would include the pair (\emph{strawberry}, \emph{vitamin C rich}) in $\mathcal{K}$. Clearly this is a heuristic strategy, which may produce non-sensical or misleading pairs. For instance, according to \textsc{MSCG}, strawberry is a \emph{low-sugar berry}, but this does not entail that strawberry has the property \emph{low-sugar} in general. However, we may expect most of the pairs that are generated with this strategy to be meaningful. A total of 8186 concept--property pairs were obtained in this way. We refer to the resulting dataset as \textsc{Prefix}. 

Finally, going beyond concept-hypernym pairs, we derive a dataset from GenericsKB \cite{DBLP:journals/corr/abs-2005-00660}, which contains generic sentences such as ``\emph{Bananas are an important food staple in the tropics}''. Due to the regular structure of such sentences, we can easily convert them into concept--property pairs; e.g.\ the aforementioned sentence would become (\emph{banana}, \emph{important food staple in the tropics}). We collect a set of 100K such pairs, by processing the sentences with the highest confidence (i.e.\ the ones which are most likely to be generic sentences) whose length is at most 7. The reason why we focus on shorter sentences is because they are more likely to capture salient information. We refer to this dataset as \textsc{GKB}.

\paragraph{Training Objective}
Given the pairs in $\mathcal{K}$, we pre-train two encoders, $\Phi_\mathsf{con}$ and $\Phi_\mathsf{prop}$, using binary cross-entropy. In particular, the loss function for a given mini-batch is defined as follows:
\begin{align*}
\mathcal{L}\, {=} &  -\sum_{(c,p)\in \mathcal{K}_{\mathsf{batch}}} \log \sigma\big(\Phi_\mathsf{con}(c)\cdot \Phi_\mathsf{prop}(p)\big)\\
& - \sum_{(c,p)\in \mathcal{N}_{\mathsf{batch}}} \log\big(1- \sigma\big(\Phi_\mathsf{con}(c)\cdot \Phi_\mathsf{prop}(p)\big)\big)
\end{align*}
Here $\mathcal{K}_{\mathsf{batch}}$ represents the subset of $\mathcal{K}$ that is in the current mini-batch. For efficiency reasons, we sample these mini-batches as follows. First, we sample a subset $\mathcal{C}_{\mathsf{batch}}$ of $K$ concepts from $\mathcal{C}$. Then, for each concept $c$ in $\mathcal{C}_{\mathsf{batch}}$ we sample one property $p\in\mathcal{P}$ such that $(c,p)\in\mathcal{K}$. Let $\mathcal{P}_{\mathsf{batch}}$ be the set of properties that are thus obtained. 
The sets of positive examples  $\mathcal{K}_{\mathsf{batch}}$ and negative examples $\mathcal{N}_{\mathsf{batch}}$ are then defined as follows:
\begin{align*}
\mathcal{K}_{\mathsf{batch}} &= (\mathcal{C}_{\mathsf{batch}} \times \mathcal{P}_{\mathsf{batch}})\cap \mathcal{K}\\
\mathcal{N}_{\mathsf{batch}} &= (\mathcal{C}_{\mathsf{batch}} \times \mathcal{P}_{\mathsf{batch}})\setminus \mathcal{K}
\end{align*}
In other words, the positive examples are the pairs from $\mathcal{K}$ that involve a concept from $\mathcal{C}_{\mathsf{batch}}$ and a property from $\mathcal{P}_{\mathsf{batch}}$. The negative examples are all the other pairs that we can form by taking a concept from $\mathcal{C}_{\mathsf{batch}}$ and a property from $\mathcal{P}_{\mathsf{batch}}$. This in-batch negative sampling strategy ensures that after encoding $|\mathcal{C}_{\mathsf{batch}}|$ concepts and $|\mathcal{P}_{\mathsf{batch}}|$ properties, we can take $|\mathcal{C}_{\mathsf{batch}}| \times |\mathcal{P}_{\mathsf{batch}}|$ training examples into account. Given that the encoders $\Phi_\mathsf{con}$ and $\Phi_\mathsf{prop}$ correspond to fine-tuned language models, and the encoding steps are thus time-consuming, in-batch negative sampling enables a significant speed-up compared to naive strategies in which positive and negative examples are sampled independently. Similar strategies are commonly used in neural information retrieval \cite{gillick-etal-2019-learning}.

\paragraph{Concept and Property Encoders}
The encoders $\Phi_{\mathsf{con}}$ and $\Phi_{\mathsf{prop}}$ correspond to fine-tuned encoder-only language models, such as BERT \cite{devlin-etal-2019-bert}.  An important design decision is how the input  to these language models is presented. For the concept encoder, the input corresponds to a string of the form ``[prefix] $c$ [suffix]'', which is usually referred to as the prompt. How this prompt is chosen often has a substantial impact on the performance of a model. For instance, language models have been reported to under-perform if the input is too short \cite{bouraoui2020inducing,jiang-etal-2020-know}. Given the importance of the choice of prompt, several strategies for automatically learning a suitable prompt have been proposed \cite{shin-etal-2020-autoprompt,liu2021gpt}. In practice, however, carefully chosen manually designed prompts often outperform such automatically learned prompts \cite{ushio-etal-2021-distilling, DBLP:journals/corr/abs-2106-13353}. For this reason, we have manually generated a number of templates and evaluated their performance on a held-out portion of the \textsc{MSCG} dataset. Based on this analysis\footnote{Details can be found in  Appendix~\ref{appendix:prompt_analysis}.}, we use the following prompt:
\begin{quote}
$\langle\textit{cls}\rangle$ [CONCEPT] means $\langle\textit{mask}\rangle\langle\textit{sep}\rangle$
\end{quote}
where $\langle\textit{cls}\rangle$, $\langle\textit{mask}\rangle$ and $\langle\textit{sep}\rangle$ are special tokens from the BERT vocabulary, while [CONCEPT] represents the concept to be modelled. The embedding of the concept is taken to be the contextualised vector of the $\langle\textit{mask}\rangle$ token, i.e.\ the representation of this token in the final layer of the language model. We use the same prompt for concepts and properties. However, note that concepts and properties are encoded using different encoders. 
Intuitively, we think of $\Phi_{\mathsf{con}}(c)$ as a representation of a prototypical instance of concept $c$, while we view $\Phi_{\mathsf{prop}}(p)$ as a representation of the property $p$ itself. This is why, even when $p=c$, we would expect $\Phi_{\mathsf{con}}(c)$  and $\Phi_{\mathsf{prop}}(c)$ to be different. Under this view, $\sigma(\Phi_{\mathsf{con}}(c) \cdot \Phi_{\mathsf{prop}}(p))$ captures the probability that a prototypical instance of $c$ would have the property $p$. In other words, by using different encoders for concepts and properties, we can capture the default nature of the pairs in $\mathcal{K}$ in a natural way.

\begin{table*}[t]
\centering
\footnotesize
\begin{tabular}{ll ccc c ccc}
\toprule
\multirow{2}{*}{\textbf{Language Model}} & \multirow{2}{*}{\textbf{\textbf{Pre-training dataset}}} & \multicolumn{3}{c}{\textbf{McRae}} && \multicolumn{3}{c}{\textbf{CSLB}} \\
 \cmidrule(lr){3-5}\cmidrule(lr){7-9}
& & \textbf{Con} & \textbf{Prop} & \textbf{C+P} & & \textbf{Con} & \textbf{Prop} & \textbf{C+P}  \\
 
 \midrule
\multicolumn{2}{l}{Random baseline}  & 26.0  & 26.5  & 26.0 && 8.6  & 8.4  & 8.6 \\
\multicolumn{2}{l}{Always true}  & 30.3  & 30.0  & 30.0 && 9.1  & 9.1  & 9.1 \\
\multicolumn{2}{l}{BERT-large sentence classifier \cite{DBLP:conf/cogsci/ForbesHC19}}    & 74 & - & - 
                       &&- &- &- \\
\multicolumn{2}{l}{Human performance \cite{DBLP:conf/cogsci/ForbesHC19}}    & 67 & - & - 
                       &&- &- &- \\
\midrule
BERT-base & No pre-training &77.7 & 30.7 & 25.2 
                    && 51.8 & 34.1 & 22.4\\ 
\midrule
BERT-base & \textsc{MSCG} & 79.9 & 46.6 & 41.6 
              && 54.0 & 36.8 & 28.9 \\
BERT-base & \textsc{Prefix} & 78.3 & 44.8 & 41.0 
                && 52.2 & 33.2 & 24.3 \\
BERT-base & \textsc{GKB}    & 79.3 & \textbf{50.7} & \textbf{46.0} 
                && 52.1 & 37.2 & 30.2 \\
BERT-base & \textsc{MSCG}+\textsc{Prefix}   & 80.2 & 47.8 & 43.2 
                                && 53.6 & 37.3 & 29.7 \\
BERT-base & \textsc{MSCG}+\textsc{GKB}  & 80.4 & 50.3 & 43.6 
                            && 54.8 & 37.1 & 28.9 \\
BERT-base & \textsc{MSCG}+\textsc{Prefix}+\textsc{GKB}  & 79.8 & 49.6 & 44.5 
                                            && 54.5 & 39.1 & 32.6 \\
 \midrule
BERT-large & No pre-training & 75.3 & 36.6 & 25.5 
                    &&54.3 & 36.4 & 17.7 \\
RoBERTa-base & No pre-training   &41.0 & 9.4 & 0.0 
                                &&38.1 & 28.7 & 9.6 \\
RoBERTa-large & No pre-training   & 73.7 & 26.9 & 9.4 
                                && 55.3 & 37.8 & 24.8 \\
 \midrule
BERT-large &\textsc{MSCG}+\textsc{Prefix}+\textsc{GKB}& \textbf{80.5} & 49.3 & 45.5 
                    &&57.7 & 41.8 & 36.4  \\
RoBERTa-base & \textsc{MSCG}+\textsc{Prefix}+\textsc{GKB}   &75.6 & 42.4 & 38.1 
                                &&49.9 & 36.4 & 24.3 \\
RoBERTa-large & \textsc{MSCG}+\textsc{Prefix}+\textsc{GKB}   &80.1 & 46.5 & 42.5 
                               && \textbf{59.0} & \textbf{42.5} & \textbf{36.0} \\

\bottomrule
\end{tabular}
\caption{Results in terms of F1 score (percentage) for commonsense property prediction.\label{tabMainResults}}
\end{table*}

\section{Experiments}
In our experiments, we primarily focus on commonsense property classification, i.e.\ predicting whether some concept has a given property. We also demonstrate the usefulness of the concept and property encoders on the task of hypernym discovery. Finally, we also present a qualitative analysis. 

\paragraph{Training Details}
We pre-train the concept and property encoders on  the datasets introduced in Section \ref{secMethodology}. We also consider variants in which these datasets are combined; e.g.\ we write \textsc{MSCG}+\textsc{Prefix} for the dataset combining the pairs from \textsc{MSCG} and \textsc{Prefix}. To pre-train our model, we construct separate validation data in the same way. In particular, for \textsc{MSCG}, we select the validation split by taking the next 10K most confident pairs from Microsoft Concept Graph (i.e.\ after removing the pairs from the \textsc{MSCG} dataset itself), and similar for the other datasets. We train the model for 100 epochs, using early stopping with a patience of 20. We use the AdamW optimizer \cite{Loshchilov2019DecoupledWD} with a learning rate of $2e{-}6$ and set the batch size to 8. We use BERT-base-uncased as our default language model \cite{devlin-etal-2019-bert}, although we have also experimented with BERT-large-uncased, RoBERTa-base and RoBERTa-large \cite{DBLP:journals/corr/abs-1907-11692}. 

\subsection{Commonsense Property Classification}
\paragraph{Datasets}
For commonsense property classification, we use the extended version of the McRae dataset \cite{mcrae2005semantic} that was introduced by \citet{DBLP:conf/cogsci/ForbesHC19}. This dataset involves a set $\mathcal{C}$ of 514 concepts and a set $\mathcal{P}$ of 50 properties. For each concept $c$ and property $p$, the dataset specifies whether $c$ has property $p$. The set $\mathcal{C}$ is split into a training set $\mathcal{C}_{\mathsf{train}}$ and a test set $\mathcal{C}_{\mathsf{test}}$\footnote{The split is available at \url{https://github.com/mbforbes/physical-commonsense}.}. During training, the models have access to the ground truth of every pair in  $\mathcal{C}_{\mathsf{train}}\times \mathcal{P}$. The models are then tested on all pairs in $\mathcal{C}_{\mathsf{test}}\times \mathcal{P}$. We report the results in terms of the  F1 score of the positive label.

As argued in the introduction, by training and testing on the same set of properties, we may not be able to faithfully test a model's ability to predict commonsense properties. For this reason, we consider an alternative setting where the set of properties is instead split into a training set $\mathcal{P}_{\mathsf{train}}$ and a test set $\mathcal{P}_{\mathsf{test}}$. During training, the model then gets access to the ground truth for the pairs in $\mathcal{C}\times \mathcal{P}_{\mathsf{train}}$, while the model is evaluated on the pairs in $\mathcal{C}\times \mathcal{P}_{\mathsf{test}}$. We use 5-fold cross-validation for this setting. Our hypothesis is that this setting will be more difficult, as it would be harder to find properties in the training data that are similar to those from the test set. However, there are nonetheless some similarities between these properties. We therefore also consider a setting in which both the concepts and properties are split into train and test fragments. The model is then trained on the pairs in $\mathcal{C}_{\mathsf{train}}\times \mathcal{P}_{\mathsf{train}}$ and evaluated on the pairs in $\mathcal{C}_{\mathsf{test}}\times \mathcal{P}_{\mathsf{test}}$. We again use a form of cross-validation. In particular, we split $\mathcal{C}$ into three folds: $\mathcal{C}_1$, $\mathcal{C}_2$ and $\mathcal{C}_3$. We similarly split $\mathcal{P}$ into three folds: $\mathcal{P}_1$, $\mathcal{P}_2$ and $\mathcal{P}_3$. In the first iteration, we train on the pairs in $(\mathcal{C}_1\cup\mathcal{C}_2)\times (\mathcal{P}_1\cup\mathcal{P}_2)$ and test on the pairs in $\mathcal{C}_3\times \mathcal{P}_3$. This process is repeated nine times (as we have three ways to choose the concept test split and three ways to choose the property test split).

We have also used the CSLB Concept Property Norms\footnote{\url{https://cslb.psychol.cam.ac.uk/propnorms}}, as a second benchmark for commonsense property classification. This dataset covers 638 concepts and 3350 properties. Similar as for McRae, the dataset indicates which concepts have which properties, but there are no standard splits. Moreover, the dataset does not explicitly contain negative examples. For this reason, for each positive example $(c,p)$, we introduce 20 negative examples by replacing $p$ with another property $p'$ (such that $(c,p')$ is not a positive example). This strategy is imperfect, as there will inevitably be some false negatives, but it should still allow us to compare the relative performance of different models. Mirroring the settings from the McRae dataset, we consider a concept-based training-test split (\textit{Con}), a property-based split (\textit{Prop}), and a setting where both concepts and properties are split into training and test sets (\textit{Con+Prop}). For consistency, we use the same cross-validation strategies as for the McRae dataset. In particular, for \textit{Con} we use a fixed split (with 90\% of the concepts used for training and 10\% for testing). For \textit{Prop}, we use 5-fold cross-validation, while for \textit{Con+Prop} we used the $3\times 3$ fold cross-validation strategy.

\paragraph{Results} 
The results for commonsense property classification are summarised in Table \ref{tabMainResults}. We include the following baselines.
First, the \emph{Random} baseline predicts the positive label with 50\% chance. Similarly, \emph{Always true} predicts the positive label in all cases.
Next, for the concept-split of the McRae dataset, we compare with the method from \citet{DBLP:conf/cogsci/ForbesHC19}, where each pair $(c,p)$ was converted into a natural language sentence. For instance, (\emph{apple}, \emph{is electrical}) is converted to the sentence ``\emph{An apple requires electricity}'', which is then fed to a BERT classifier. Due to its manual nature, this method cannot be applied to new properties. We also include the estimate of human performance that was reported by \citet{DBLP:conf/cogsci/ForbesHC19}. Finally, we consider a variant of our model which is directly trained on the McRae and CSLB training data, without the pre-training step.

The next set of results compare the performance of the different pre-training datasets. For these results, all models were initialised with BERT-base. We can clearly see that the pre-trained bi-encoder models outperform the variant without pre-training in nearly all settings (with the results for \textsc{Prefix} on the CSLB property-split the only exception). This confirms our hypothesis that Microsoft Concept Graph and GenericsKB capture useful information about commonsense properties. Comparing the different corpora, \textsc{Prefix} achieves the weakest results, which can be explained by the much smaller size of this dataset. However, combining \textsc{Prefix}+\textsc{MSCG} outperforms \textsc{MSCG} in all but one case. Furthermore, as expected, the property-split (\textit{Prop}) is considerably harder than the standard concept-split (\textit{Con}), with the \textit{C+P} setting being even harder. In fact, for the latter setting, the BERT-base model without pre-training cannot outperform the random classifier for McRae. Note that for CSLB, outperforming the random classifier is easier, given that more training data is available for that dataset. Crucially, while the best baselines only slightly underperform the pre-trained models for the concept-split, much larger differences are seen for the other splits. Overall, these findings confirm our hypothesis that predicting commonsense properties remains a highly challenging problem.


Finally, the table also shows results for some other language models. While the large models generally outperform their base counterparts, the differences are relatively small, and the improvements are not consistent. This finding is in accordance with the conclusion from \citet{DBLP:journals/corr/abs-2111-00607} that even large language models are limited in the amount of commonsense knowledge they capture, and in particular that finding the right pre-training task is crucial. The RoBERTa results without pre-training are particularly disappointing, with learning failing completely in some cases. Even with the pre-training datasets, BERT-base outperforms RoBERTa base, and BERT-large outperforms RoBERTa-large.

\begin{table}[t]
\centering
\footnotesize
\begin{tabular}{lccc}
\toprule
& \textbf{Con} & \textbf{Prop} & \textbf{C+P} \\
\midrule
Skip-gram ($k=1$) & 70.8 & 25.0 & 17.5 \\
Skip-gram ($k=3$) & 53.4 & 9.5 & 5.7 \\
GloVe ($k=1$) & 68.8 & 20.3 & 21.7 \\
GloVe ($k=3$) & 51.4 & 6.8 & 4.9 \\
BERT-base ($k=1$) & \textbf{72.0} & \textbf{28.2} & \textbf{27.0} \\
BERT-base ($k=3$) & 55.6 & 14.6 & 19.1 \\
\bottomrule
\end{tabular}
\caption{Evaluation of a nearest neighbour strategy for the McRae dataset (F1 score percentage). 
\label{tabNNClassifier}}
\end{table}

\paragraph{Analysis}
As we have argued, models can perform well on the \textit{Con} setting by simply transferring knowledge about similar concepts from the training data. This is analysed in more detail in Table \ref{tabNNClassifier}, which shows the performance of a nearest neighbour classifier. To classify a test pair $(c,p)$ we find the $k$ concepts $c_1,...,c_k$ from the training split that are most similar to $c$ in terms of cosine similarity. Then we predict the positive label for $(c,p)$ if the majority of $(c_1,p),...,(c_k,p)$ are assigned the positive label. We test this approach for $k=1$ and $k=3$, using embeddings from GloVe \cite{pennington-etal-2014-glove} and Skip-gram\footnote{We used the 300 dimensional Skip-gram vectors trained on Google News and GloVe vectors trained on Common Crawl, available from \url{https://radimrehurek.com/gensim/models/word2vec.html}.} \cite{mikolov-etal-2013-linguistic}, and using the embeddings predicted by our BERT-base encoder pre-trained on \textsc{MSCG}+\textsc{Prefix}+\textsc{GKB}.
For the \textit{Prop} setting, we similarly predict the label of $(c,p)$ based on the labels of the training pairs $(c,p_1),...,(c,p_k)$, with $p_1,...,p_k$ the $k$ properties from the training data that are most similar to $p$. Finally, for $\textit{C+P}$, we predict the majority label among the training pairs $(c_i,p_j)$ with $i,j\in\{1,...,k\}$, where $c_1,...,c_k$ are the training concepts most similar to $c$ and $p_1,...,p_k$ are the training concepts most similar to $p$. 
The results in Table \ref{tabNNClassifier} clearly support our hypothesis about the concept-split. In particular, the nearest neighbour classifier is highly effective for the concept-split (for $k=1$), outperforming the estimate of human performance from \citet{DBLP:conf/cogsci/ForbesHC19} for all embedding types, and approaching the performance of the language models without our pre-training task. In contrast, for the \textit{Prop} and \textit{C+P} settings, the nearest neighbour classifier performs, at best, similarly to the random classifier. 



\begin{table}[t]
\footnotesize
\centering
\begin{tabular}{l@{\hspace{7pt}}l@{\hspace{7pt}}ccc}
\toprule
& & MAP & MRR & P@5 \\
\midrule
\parbox[t]{2mm}{\multirow{5}{*}{\rotatebox[origin=c]{90}{\textbf{General}}}}  & APSyn & 1.7  & 3.7   & 1.7  \\
& balAPInc & 1.7   & 3.9   & 1.7  \\
& SLQS & 0.7  & 1.7  & 0.7  \\
& Apollo & 2.7   & 6.1   & 2.8  \\
& Ours  &  \textbf{3.8}  & \textbf{7.0}   & \textbf{3.1}  \\
\midrule
\parbox[t]{2mm}{\multirow{5}{*}{\rotatebox[origin=c]{90}{\textbf{Music}}}} & APSyn & 1.1  & 2.6   & 1.3  \\
& balAPInc & 1.4   & 3.6   & 1.6  \\
& SLQS & 0.6  & 1.3   & 0.7  \\
& ADAPT & 1.9   & \textbf{5.3}  & 1.9  \\
& Ours  & \textbf{2.3}   & 5.1  & \textbf{2.6}  \\
\midrule
\parbox[t]{2mm}{\multirow{5}{*}{\rotatebox[origin=c]{90}{\textbf{Medical}}}} & APSyn & 0.7   & 1.4   & 0.7  \\
& balAPInc & 0.9 & 2.1  & 1.1  \\
& SLQS & 0.3   & 0.7   & 0.3  \\
& ADAPT & \textbf{8.1}   & \textbf{20.6}   & \textbf{8.3}  \\
& Ours  & 4.0   & 9.0   & 3.9  \\
\bottomrule
\end{tabular}
\caption{Result of the hypernym discovery experiment. 
\label{TabHypDiscResults}
}
\end{table}

\subsection{Hypernym Discovery}
Given an input word (e.g.\ \textit{cat}), the aim of the  \textit{hypernym discovery} task is to retrieve a set of valid hypernyms (e.g.\ \textit{animal}, \textit{mammal}, \textit{feline}, etc.). 
We use this task to analyse the quality of the pre-trained concept and property encoders when used without any fine-tuning on task-specific training data. We  use the data from the SemEval 2018 Hypernym Discovery task \cite{semeval2018task9}, focusing on the concept-only split (i.e.\ without considering named entities). There are three variants of this task: an open-domain setting (referred to as \emph{general}) and two domain-specific settings, focusing on the \emph{music} and \emph{medical} domains. Each variant is associated with a large vocabulary of candidate terms, consisting of 218,753 terms for \emph{general}, 69,118 terms for \emph{music} and 93,888 terms for \emph{medicine}. To solve this task, each word from the vocabulary is encoded using $\Phi_{\mathsf{prop}}$. We then use maximum inner product search to efficiently find those words $w$ from the vocabulary that maximise $\Phi_{\mathsf{con}}(t)\cdot \Phi_{\mathsf{prop}}(w)$ for a given target word $t$. From the retrieved list of words, we remove those that contain the term $t$ itself and those that end with an adjective. For this experiment, we use BERT-large encoders pre-trained on \textsc{MSCG}+\textsc{Prefix}+\textsc{GKB}. We compare our method with the following baselines for this task: APSyn \cite{santus2016unsupervised},  balAPInc \cite{kotlerman2010directional}, SLQS \cite{santus2014chasing}, ADAPT \cite{maldonado-klubicka-2018-adapt} and Apollo \cite{onofrei-etal-2018-apollo}. We report the published results from the SemEval task \citet{semeval2018task9} (where ADAPT only participated in the \emph{general} setting and Apollo only participated in the \emph{music} and \emph{medical} settings). The latter systems achieved the best performance among the unsupervised methods\footnote{The hypernym discovery datasets are strongly biased in which hypernyms were preferred by the annotators. Such biases can only be learned from the task-specific training data, which is why we do not compare with supervised methods.}. Following \citet{semeval2018task9}, we report Mean Average Precision (MAP), Mean Reciprocal Rank (MRR), and Precision at $5$ ($P@5$), in percentage terms. Table \ref{TabHypDiscResults} shows that our method outperforms all baselines for \emph{General}, performs similar to ADAPT for \emph{Music} and worse than ADAPT for \emph{Medical}. This is remarkable, given that our method was not designed or tuned for this task. The under-performance on \emph{Medical} can be explained by the lack of training examples from this domain in the pre-training data. As can be observed, the results for all models are low. An error analysis, presented below, revealed that this is largely due to the fact that many correct hypernyms are not included in the ground truth.

\paragraph{Error Analysis}
Table \ref{tabHypernymDiscErrorAnalysis} shows some of the predictions of our model for the \emph{General} setting of the hypernym discovery task. The first set of results shows examples where many of the predicted hypernyms are intuitively correct. However, only few of these hypernyms are covered by the ground truth; ground truth predictions are shown in bold. This illustrates the rather noisy nature of the dataset, and serves as an explanation for the low overall F1 score of the different unsupervised models. The second set of results in Table \ref{tabHypernymDiscErrorAnalysis} covers cases where most of the predictions are incorrect. In some cases, e.g.\ for \emph{children}, the model predicts semantic properties rather than hypernyms, which shows that simply filtering predictions that end with an adjective is not always sufficient. The case of \emph{broiler chicken} shows that the model sometimes predicts terms that are semantically related, but which are clearly not hypernyms (nor semantic attributes). As a variant of this observation, the case of \emph{sigma} shows that the model sometimes tends to predict co-hyponyms.

\begin{table*}[h!]
\centering
\footnotesize
\begin{tabular}{ll}
\toprule
\textbf{Hyponym} & \textbf{Top-5 Predicted Hypernyms} \\
\midrule
liberty & principle, notion, ideal, universal value, humanitas \\
longbow & hunting weapon, \textbf{weapon}, bow and arrow, wieldy, choptank\\
wine & \textbf{drink}, \textbf{beverage}, liquidity, alcoholic beverage, drinking alcohol\\
manslaughter & culpable homicide, murder charge, offence, justifiable homicide, first-degree murder\\
shopping & chore, specific activity, everyday, simple interest, pursuit\\
running & aerobic, cardio, endurance training, aerobic exercise, sport\\
computer industry & sector, sunrise industry, growth industry, field of operation, game industry\\
learner & understander, student, realizer, know-all, nonjoinder\\
snow & weather condition, weather, cold weather, bad weather, wet-weather\\
bounty hunter & vigilante, hired gun, bandit, bondman, trail boss\\
metre & \textbf{unit of length}, unit of measure, measuring unit, quantity unit, derived unit\\
hero & protagonist, archetype, archetypic, personage, literaty character\\
website & resource, e-resource, information source, \textbf{medium}, source\\
violin & \textbf{string instrument}, \textbf{musical instrument}, second fiddle, \textbf{bowed instrument}, \textbf{stringed instrument}\\
\midrule
arms & head and shoulders, legs, straighten, stiffen, bare bones\\
cooking ingredient & composition, culinary, adjunct, importune, condiment\\
children & learn, memorize, make fun, come to life, lose track\\
broiler chicken & chicken cordon bleu, chicken stock, hot chicken, kung pao chicken, chicken broth\\
observation & qualitative, empirical research, qualitative analysis, data collection, qualitative research\\
sigma & lambda, upsilon, fraternity, epsilon, alpha and omega\\
apartment & tenantless, adjacent, low-rent, homeplace, residential building\\
wetsuit & drysuit, nonsuit, life-jacket, diving equipment, diving suit\\
yesterday & thisday, tomorrow, timea, timeless, evermore\\
taxi & off-license, car rental, bus service, bike rental, cab fare\\
\bottomrule
\end{tabular}
\caption{Error analysis for hypernym discovery on the general dataset. Correctly predicted hypernyms are shown in bold.\label{tabHypernymDiscErrorAnalysis}}
\end{table*}

\subsection{Qualitative Analysis}
As a qualitative analysis, we use our pre-trained models to predict which properties are associated with a given concept. We consider the set of all properties that appear at least 10 times in an extended version of the 
\textsc{Prefix}+\textsc{GKB} dataset\footnote{This extended dataset involves 500K pairs from Microsoft Concept Graph and 500K sentences from GenericsKB; analysis about this extended dataset is provided in Appendix~\ref{appendix:size_pretrain_corpus}.}, leading to a set of 5223 candidate properties. We again use maximum inner product search to efficiently identify the properties whose embeddings are closest to the concept embedding $\phi_{\mathsf{con}}(c)$. Table \ref{tabQualitativeAnalysis_large} shows the seven nearest properties for a number of selected concepts, where we used BERT-base pre-trained on \textsc{MSCG}+\textsc{Prefix}+\textsc{GKB}. Specifically, the table first revisits the examples from Table \ref{tabPropertiesIntro}. Subsequently, the table lists physical concepts, for which we expected predicting properties to be easier, and abstract concepts, for which we expected the task to be harder. Finally, we included adjectives to explore whether our model can be used for learning property entailment.

\begin{table*}[h!]
\centering
\footnotesize
\begin{tabular}{l@{\hspace{5pt}}p{380pt}}
\toprule
\textbf{Concept} & \textbf{Predicted properties} \\
\midrule
banana  & food, fruit, fresh, plant, edible, tropical, commercially important \\
lion & animal, mammal, wildcat, carnivore, species, very territorial, mammalian\\
airplane & vehicle, aircraft, stationary, application, object, military vehicle, automotive\\
\midrule
straw  & material, combustible, porous, stuff, fibrous, located in wood, has sections \\
ice  & cold, has temperature, has surfaces, located in freezers, has density, authorization, albums \\
yacht  & boat, vehicle, vessel, recreational, ship, expensive, aircraft\\
coffee & beverages, drinks, beverage, drink, liquid, liquids, located in supermarkets \\
steel  & material, non-ferrous, non ferrous, rigid, product, industrial, heavy \\
fire  & causes burns, creates heat, produces heat, causes damage, can have effects, generates heat, produce crops \\
beer  & beverage, drink, alcoholic, liquor, liquid, beverages, drinks\\
\midrule
democracy  & principle, idea, democratic, ideology, concept, morality, value, moral \\
disappointment  & negative, feeling, emotion, emotional, feelings, positive, depression \\
promotion  & marketing, achievement, activity, corporate, factor, acts, activities\\
celebration  & event, festivity, occasion, social events, parties, events, activities \\
forgiveness  & moral, value, love, virtue, emotion, benign, principle\\
lawyer  & professional, adult, allied, profession, consultant, closely related, expert\\
\midrule
stressful  & situation, factor, emotional, difficult, unexpected, uncomfortable, traumatic \\
poisonous & poison, harmless, harmful, dangerous, toxin, aggressive, sharp\\
sugary  & dessert, taste, food, delicious, chocolate, frozen dessert, candy \\
rewarding  & activities, clocks, happiness, treatments, approval, actions, human activities \\
modern & style, genre, contemporary, fashion, broad, musical style, english \\
alcoholic & alcoholic, liquor, drink, beverage, mixed, alcohol, addictive, aggressive\\
\bottomrule
\end{tabular}
\caption{Qualitative analysis, showing the top neighbours of the embeddings of selected concepts.\label{tabQualitativeAnalysis_large}}
\end{table*}

The results contain a mixture of hypernyms and semantic attributes, which is a reflection of how the model was trained. For physical concepts, the results are generally meaningful, with a few exceptions. For instance, \emph{military vehicle} is incorrectly listed as a hypernym of \emph{airplane}. 
Regarding the abstract concepts, the top predictions are mostly meaningful, but we can also see terms that are semantically related but are neither hypernyms nor semantic attributes; e.g.\ we see \emph{parties} as a property of \emph{celebration}. Finally, for the adjectives, we see several instances where the entailment direction is reversed, for instance when \emph{dessert} is mentioned as a property of \emph{sugary}.

\section{Conclusions}
We studied the problem of modelling the commonsense properties of concepts. We argued that the standard evaluation setting does not faithfully assess the extent to which models capture knowledge about commonsense properties, and proposed two new evaluation settings. These new settings were found to be highly challenging for language models, with performance being close to random. We furthermore found that pre-training a bi-encoder model on hypernymy data or generic sentences can lead to substantial performance gains. However, there remains a lot of room for further improvements, which will likely require novel insights. 

\section*{Acknowledgments}
This work has been supported by EPSRC grant EP/V025961/1. We acknowledge the support of the Supercomputing Wales project, which is part-funded by the European Regional Development Fund (ERDF) via Welsh Government.

\bibliography{main}
\bibliographystyle{acl_natbib}

\appendix

\section{Prompt Analysis}
\label{appendix:prompt_analysis}

Previous work has found that the prompt which is used can have a material impact on the performance of BERT-based encoders \cite{bouraoui2020inducing,jiang-etal-2020-know,shin-etal-2020-autoprompt,liu2021gpt,ushio-etal-2021-distilling,DBLP:journals/corr/abs-2106-13353}. To analyse the impact of the prompt in our setting, and make a suitable choice, we experimented with a number of different, manually chosen prompts. 
For these experiments, we used the most confident 11,000 concept-property pairs of the \textsc{MSCG} dataset for training, and the next 1200 concept-property pairs for tuning. The batch size is set to 8. We used the AdamW optimizer and learning rate $2e{-}6$, using early stopping with a patience of 20. The results in Table \ref{tabPromptAnalysis} are reported in terms of the F1 score (percentage) of the positive label. For the first two results in the table, a different prompt was used for the concept and property encoders. The property prompts corresponding to these two configurations are (not shown in the table):
\begin{itemize}
\item $\langle\textit{cls}\rangle$ Property: [CONCEPT] $\langle\textit{sep}\rangle$
\item $\langle\textit{cls}\rangle$ Yesterday, I saw a thing which is [PROPERTY] $\langle\textit{sep}\rangle$
\end{itemize}
For the first six configurations in the table, we use the average of the embeddings of all tokens, in the final layer of the BERT-base model, as the embedding of the concept and property. For the remaining seven configurations, we use the embedding of the $\langle\textit{mask}\rangle$ token in the final layer instead. The results show that many of the prompts lead to a relatively similar performance, as long as the prompt is sensible. The example with the nine mask tokens (Prompt 5) show that without a semantically informative prompt the performance drops somewhat. A similar observation can be made for the prompt about the spaceship (Prompt 10). Earlier work has suggested that longer prompts tend to perform better. To some extent this is confirmed by our results. For instance, Prompt 12 outperforms the similar but shorter Prompts 9 and 11, although Prompt 13, which is an even longer variant, performs worse. Moreover, we can see that some of the shortest prompts nonetheless perform well. All things being equal, having a shorter prompt is desirable, as it means we can use larger batch sizes and faster training. For this reason, we have decided, based on these results, to use Prompt 7, whose performance is close to that of the best-performing prompt, despite also being one of the shortest ones.

\begin{table*}[t]
\footnotesize
\begin{tabular}{l@{\hspace{5pt}}lc}
\toprule
& \multicolumn{1}{c}{\textbf{Prompt}} & \textbf{F1} \\
\midrule
1.& $\langle\textit{cls}\rangle$ Concept: [CONCEPT] $\langle\textit{sep}\rangle$  & 85.6\\
2.& $\langle\textit{cls}\rangle$ Yesterday, I saw another [CONCEPT] $\langle\textit{sep}\rangle$   & 86.1\\
\midrule
3.& $\langle\textit{cls}\rangle$ The notion we are modelling is [CONCEPT] $\langle\textit{sep}\rangle$ & 86.7\\
4.& $\langle\textit{cls}\rangle$ The notion we are modelling: [CONCEPT] $\langle\textit{sep}\rangle$ & 87.3\\
5.& $\langle\textit{cls}\rangle\,\langle\textit{mask}\rangle\,\langle\textit{mask}\rangle\,\langle\textit{mask}\rangle\,\langle\textit{mask}\rangle\,\langle\textit{mask}\rangle$  [CONCEPT] $\langle\textit{mask}\rangle\,\langle\textit{mask}\rangle\,\langle\textit{mask}\rangle\,\langle\textit{mask}\rangle\,\langle\textit{sep}\rangle$  & 84.8\\
6.& $\langle\textit{cls}\rangle$ The notion we are modelling is called CONCEPT $\langle\textit{sep}\rangle$  & 86.0\\
\midrule
7.& $\langle\textit{cls}\rangle$ CONCEPT means $\langle\textit{mask}\rangle\,\langle\textit{sep}\rangle$  & 87.1\\
8.& $\langle\textit{cls}\rangle$ CONCEPT  $\langle\textit{sep}\rangle\,\langle\textit{mask}\rangle\,\langle\textit{sep}\rangle$  & 86.6\\
9.& $\langle\textit{cls}\rangle$ The notion we are modelling is CONCEPT  $\langle\textit{sep}\rangle\,\langle\textit{mask}\rangle\,\langle\textit{sep}\rangle$  & 86.8\\
10.& $\langle\textit{cls}\rangle$ The spaceship we are modelling is CONCEPT  $\langle\textit{sep}\rangle\,\langle\textit{mask}\rangle\,\langle\textit{sep}\rangle$  & 85.8\\
11.& $\langle\textit{cls}\rangle$ We are modelling CONCEPT  $\langle\textit{sep}\rangle\,\langle\textit{mask}\rangle\,\langle\textit{sep}\rangle$  & 86.4\\
12.& $\langle\textit{cls}\rangle$ The notion we are modelling this morning is CONCEPT  $\langle\textit{sep}\rangle\,\langle\textit{mask}\rangle\,\langle\textit{sep}\rangle$  & 87.0\\
13.& $\langle\textit{cls}\rangle$ As I have mentioned earlier, the notion we are modelling this morning is CONCEPT  $\langle\textit{sep}\rangle\,\langle\textit{mask}\rangle\,\langle\textit{sep}\rangle$  & 86.3\\
\bottomrule
\end{tabular}
\caption{Performance of different prompts on a held-out portion of the \textsc{MSCG} dataset, in terms of F1-score percentage. BERT-base was used as the language model in these experiments.\label{tabPromptAnalysis}}
\end{table*}

\section{Size of the Pre-Training Corpus}
\label{appendix:size_pretrain_corpus}

The \textsc{MSCG} corpus was obtained by taking the 100K pairs from Microsoft Concept Graph \cite{DBLP:journals/dint/JiWSZWY19} with the highest confidence. Similarly, \textsc{GKB} was obtained by taking the 100K sentences with the highest confidence in GenericsKB \cite{DBLP:journals/corr/abs-2005-00660}. This choice represents a trade-off: choosing more pairs would increase the overall amount of training data, which could improve the performance of the encoders. However, this would also mean including less reliable pairs, which might have a negative effect. In particular, both Microsoft Concept Graph and GenericsKB have been extracted from text corpora. In both cases, it can be clearly observed that the pairs/sentences with the lowest confidence are often rather noisy. To analyse this trade-off, Table \ref{tabExp500k} shows the results of an experiment where we used the top 500K pairs in \textsc{MSCG} and the 500K most confidence sentences in GenericsKB. Similarly, \textsc{Prefix} was derived from the larger \textsc{MSCG} dataset for these experiments. The results show a small improvement for \textsc{MSCG}. However, for the \textit{Prop} and \textit{C+P} settings, the \textsc{GKB} results are actually worse for the 500K setting. These results suggest that the optimal setting might use more than 100K pairs from Microsoft Concept Graph, but fewer than 500K sentences from GenericsKB. However, the results also show that any performance gains arising from optimising the selection of the pre-training data are likely to be small.

\begin{table}[t]
\centering
\footnotesize
\begin{tabular}{llccc}
\toprule
& & \textbf{Con} & \textbf{Prop} & \textbf{C+P} \\
\midrule
\parbox[t]{2mm}{\multirow{6}{*}{\rotatebox[origin=c]{90}{\textbf{500K}}}} & \textsc{MSCG} & 80.1 & 48.6 & 42.8 \\
& \textsc{Prefix} & 78.1 & 45.0 & 41.7 \\
& \textsc{GKB}    & 80.6 & 48.8 & 43.7 \\
& \textsc{MSCG}+\textsc{Prefix}   & 79.8 & 49.1 & 43.4 \\
& \textsc{MSCG}+\textsc{GKB}  & \textbf{80.7} & 48.8 & 41.5 \\
& \textsc{MSCG}+\textsc{Prefix}+\textsc{GKB} & 80.3 & 47.5 & 41.0 \\
\midrule
\parbox[t]{2mm}{\multirow{6}{*}{\rotatebox[origin=c]{90}{\textbf{100K}}}} & \textsc{MSCG} & 79.9 & 46.6 & 41.6 \\
& \textsc{Prefix} & 78.3 & 44.8 & 41.0 \\
& \textsc{GKB}    & 79.3 & \textbf{50.7} & \textbf{46.0} \\
& \textsc{MSCG}+\textsc{Prefix}   & 80.2 & 47.8 & 43.2 \\
& \textsc{MSCG}+\textsc{GKB}  & 80.4 & 50.3 & 43.6 \\
& \textsc{MSCG}+\textsc{Prefix}+\textsc{GKB} & 79.8 & 49.6 & 44.5 \\
\bottomrule
\end{tabular}
\caption{Evaluation on the McRae dataset of a variant in which 500K pairs from Microsoft Concept Graph and GenericsKB were used. Results are reported in terms of F1 score percentage. BERT-base was used as the language model in these experiments.\label{tabExp500k}}
\end{table}

\end{document}